\documentclass[10pt,twocolumn,letterpaper]{article}

\usepackage{iccv}
\usepackage{times}
\usepackage{epsfig}
\usepackage{graphicx}
\usepackage{amsmath}
\usepackage{amssymb}

\usepackage[breaklinks=true,bookmarks=false]{hyperref}

\iccvfinalcopy 

\def\iccvPaperID{****} 

\ificcvfinal\fi

\begin{document}

\title{CurlingNet: Compositional Learning between Images and Text for Fashion IQ Data}

\author{Youngjae Yu\\
RippleAI \& SNU\\
Seoul, Korea \\
{\tt\small yj.yu@rippleai.co}
\and
Seunghwan Lee\\
RippleAI\\
Seoul, Korea\\
{\tt\small seunghwan@rippleai.co}
\and
Yuncheol Choi\\
RippleAI \\
Seoul, Korea\\
{\tt\small ycchoi@rippleai.co}
\and
GunheeKim\\
RippleAI \& SNU \\
Seoul, Korea\\
{\tt\small gunhee@snu.ac.kr}
}

\maketitle
\ificcvfinal\thispagestyle{empty}\fi

\begin{abstract}
We present an approach named \textit{CurlingNet} that can measure the semantic distance of composition of image-text embedding. 
In order to learn an effective image-text composition for the data in the fashion domain, our model proposes two key components as follows.
First, the \textit{Delivery} makes the transition of a source image in an embedding space. 
Second, the \textit{Sweeping} emphasizes query-related components of fashion images in the embedding space.
We utilize a channel-wise gating mechanism to make it possible.
Our single model outperforms previous state-of-the-art image-text composition models including TIRG~\cite{vo-cvpr-2019} and FiLM~\cite{perez-AAAI-2018}. 
We participate in the first fashion-IQ challenge~\cite{fashioniq-arxiv-2019} in ICCV 2019, for which ensemble of our model achieves one of the best performances. 

\end{abstract}

\section{Introduction}

Controllable image search based on user input (\eg natural language query and category filter) is an important topic of research to technically study the multimodal embedding and practically promote user convenience of search.
It can provide a significant contribution for developing an interactive conversational image search systems.

\begin{figure}[t]
\centering
\includegraphics[trim=0.0cm 0.0cm 0cm 0.0cm,clip,width=0.47\textwidth]{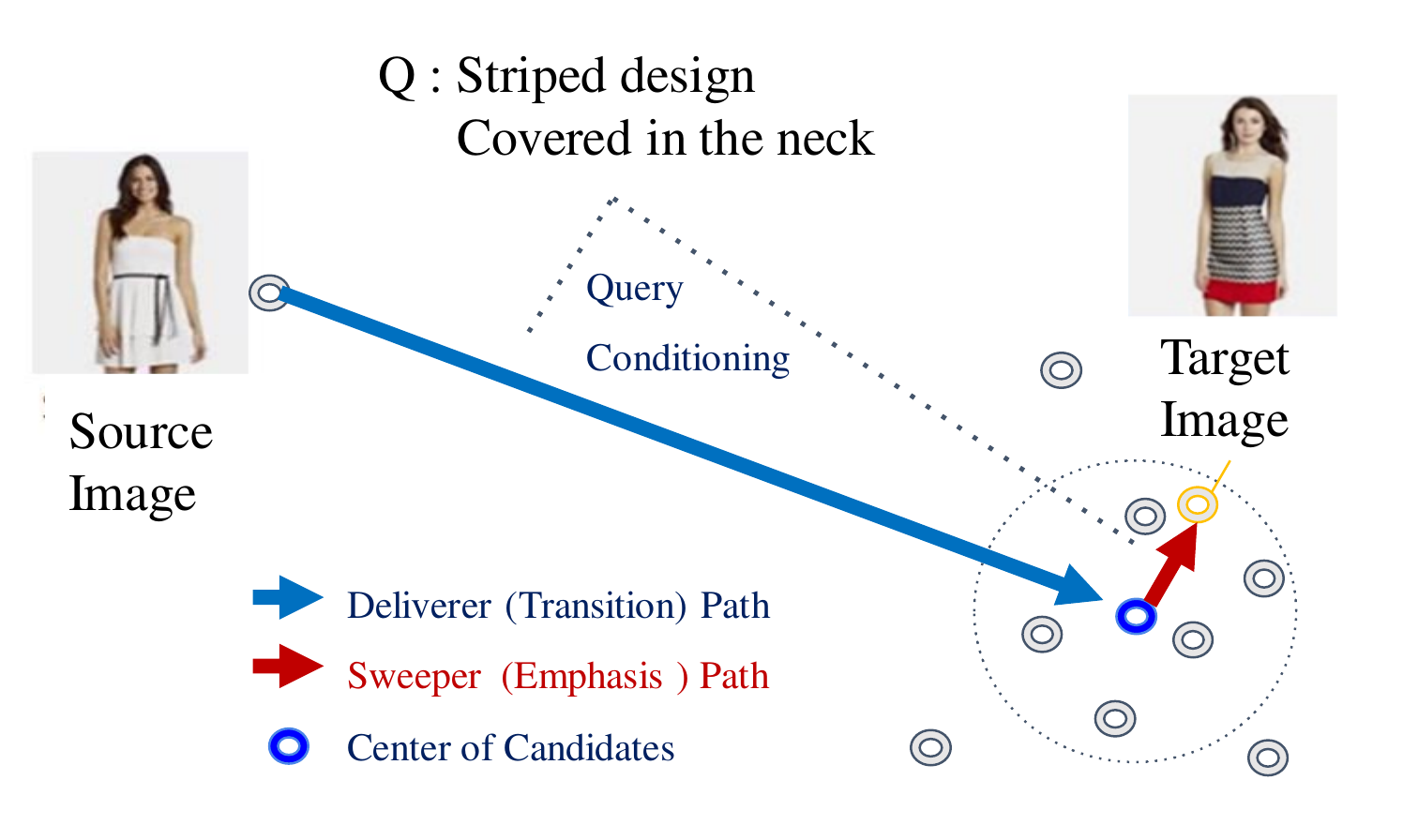}
\caption{
The key intuition of the CurlingNet model. 
Given a triplet of (Source image, Language query, Target image), 
Our model effectively encodes the differential relationship between the two images.
}
\label{fig:keyidea}
\end{figure}

In this report, we aim at proposing an approach that can measure the semantic similarity between the composition embedding of images and text, and applying it to tackle controllable multimodal image retrieval for fashion data.
Our model, named \textit{Curling Network}, consists of two main components, \textit{Delivery} and \textit{Sweeping} as shown in Figure~\ref{fig:keyidea}.
The delivery path guides the image embedding into the imaginary center of candidate images, which will be an ideal target point.
The sweeping path learns to make the distance closer to the target images with the same properties as described in the query text.
That is,  the sweeping component moves the target embedding point to emphasize key attributes in the query text (\eg striped, covered in the neck). 

We summarize the contributions of this work as follows.

\begin{enumerate}
  \vspace{-3pt}\item We propose the CurlingNet, which can measure the semantic differential relationship between images with respect to a query text. 
  Compare to existing models such as TIRG~\cite{vo-cvpr-2019}, our model not only learns the  image-text composition features for finding the most suitable target data, but also focuses on the queried attributes in target data for better ranking.

\vspace{-3pt}\item To validate our proposed model, we participate in Fashion-IQ challenge and our ensemble model achieves the second place on the leaderboard. 
\end{enumerate}

\section{Approach}

\subsection{Preprocessing}

\textbf{Fashion Attribute Experts.}
We use the image feature encoded by pre-trained image CNNs (\eg ResNet-152~\cite{he-arxiv-2015}, DenseNet-169 ~\cite{huang-cvpr-2017}) as a basic expert.
In addition, we treat the Fashion IQ attributes of each category as another expert.
as a result, a total of seven experts are used to encode the source and target images. 
We embedded the attributes using the pretrained word2vec~\cite{Tomas-nips-2013}.
The number of attributes per data ranges from 0 to 19, so we pool them with a learnable pooling method, NetVLAD~\cite{arand-cvpr-2016}. 
It captures information about the statistics of local descriptors aggregated over the attributes. 
Then, six expert features are obtained by combining image features and the pooled attributes. 
These features are supposed to represent the attribute information in the image.
We use Context Gating~\cite{miech-arxiv-2017} to combine image features and attributes.
Finally, we obtain the encoding of the experts from Collaborative Expert (CE) gating~\cite{Liu-bmvc-2019}.

\textbf{Text Encoding.}
For text encoding, we use the three-level encoding method \cite{dual-cvpr-2019}: (1) global encoding by mean pooling, 	(2) temporal-aware encoding by biGRU and (3) Locally-enhanced encoding by biGRU-CNN.
We concatenate the encodings of all levels to make the final text feature, so that the text encoding includes multi-level information. 

\begin{figure}[t]
\centering
\includegraphics[trim=0.0cm 0.0cm 0cm 0.0cm,clip,width=0.47\textwidth]{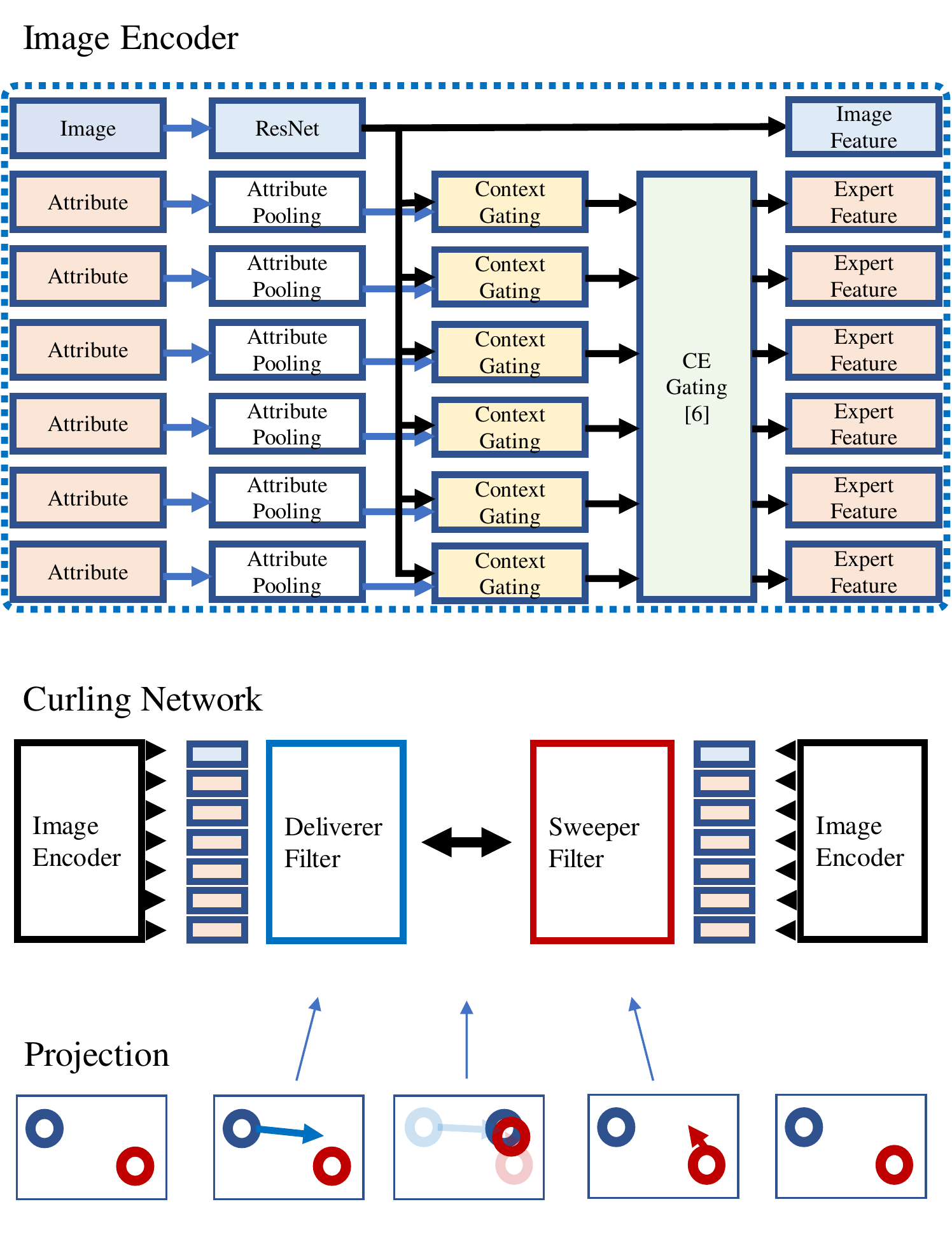}
\caption{
The architecture of the image encoder using Collaborative Experts~\cite{Liu-bmvc-2019}.
}
\label{fig:curling}
\end{figure}

\subsection{Transition Filter Networks}

\begin{figure}[t]
\centering
\includegraphics[trim=0.0cm 0.0cm 0cm 0.0cm,clip,width=0.47\textwidth]{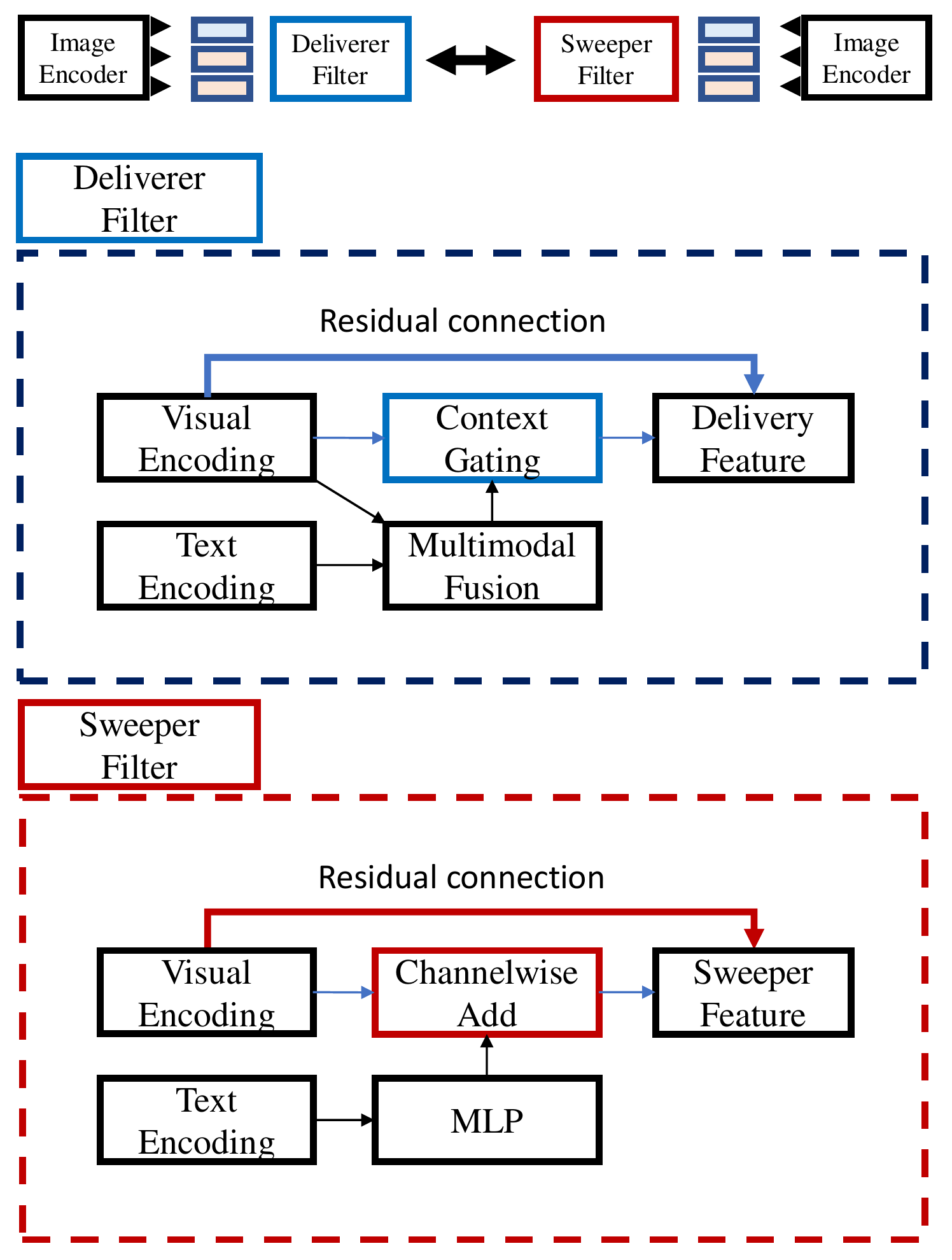}
\caption{
Schematics for the Deliverer filter and the Sweeper filter. The dense projection layer is omitted. Best viewed in color.
}
\label{fig:filter_detail}
\end{figure}

In Figure \ref{fig:keyidea}, suppose that the text query is \textit{less covered on the neck and is fully patterned}  for a given source image.
However, there can be many other differences beyond these two query requests between the true source and taget image pair,
such as shorter skirts, sticking to the body, and red and blue.
In other words, the target image cannot be fully specified by text query; instead, only some center points of various valid targets can be specified with one source and text query. 
Therefore, we need two types of networks.
The first one delivers the source image to the candidate cluster according to a given query in an embedding space (\ie TIRG ~\cite{vo-cvpr-2019}).
The second network checks the attributes highlighted in the query and learn the path from the center of valid target candidates to the true target image. 
We design two networks named \textit{Delivery filters} and \textit{Sweeping filters},
whose descriptions are shown in Figure~\ref{fig:filter_detail}.

\textbf{Multimodal Fusion Layers.}
Image expert features and text encodings are taken into the multimodal fusion layer.
We use concatenation of these multimodal inputs and their output of the fusion function, for which we use a simple Hadamard product so that each of the two inputs is projected into a semantic space with the same dimension size.
\begin{align}
\label{eq:fusion}
   \text{FusionLayer}(A,B) = [A ;B ; A \odot B] \nonumber
\end{align}%
For final ensemble, we use several variants of fusion functions such as MUTAN~\cite{mutan-iccv-2017} and MCB~\cite{mcb-emnlp-2016}. 

\textbf{The Delivery Filter.}
We use the subsequent context gating to manipulate the vector transition after the multimodal fusion pooling. 
It is inspired by TIRG~\cite{vo-cvpr-2019} as an image-text composition technique that is successfully used for image-text retrieval tasks.
We add 3 dense projection layers with BatchNorm~\cite{Sergey-icml-2015} prior to the sigmoid gating function.

\textbf{The Sweeping Filter.}
We use the channel-wise addition on target image encoding with visual-difference text query.
As described in 2.1, we compute text feature and fusion with target expert features using ~\cite{mutan-iccv-2017}. Then we obtain an adjusted feature after simple addition. The adjusted feature is summed up with the residual connection to target expert features.


\textbf{External Dataset.}
We use fashion-200K~\cite{han-iccv-2017} and fashion-gen~\cite{arxiv-fashiongen-2018} dataset as pretraining sources for Only-Resnet-Encoder model.
For fashion-gen dataset, we only use four category, \textit{SHIRTS}, \textit{SWEATERS}, \textit{TOPS}, and \textit{DRESSES}.
Following fashion-200K post-processing strategy in \cite{vo-cvpr-2019}, we create (source image, query text, target image) triplets from both fashion-200K and fashion-gen datasets.

\subsection{Training and Evaluation}
We use the additive margin softmax as our loss function~\cite{amloss-arxiv-2019}.
Each training batch consists of $L$ triplets of (source image, query text, target image).
We use batch shuffling in every training epoch.
The total similarity between the source and target image combined with query text is obtained by the weighted sum of normalized dot products calculated by each expert. 
The weight for weighted sums can be learned from the text encoding. For data with no or only parts of six attributes, the weights for the missing attributes are set to 0 and renormailized so that they would not be used for the similarity calculation~ \cite{Liu-bmvc-2019}.
We use the Adam~\cite{kingma-iclr-2015} optimizer. 

We set the Initial learning rate as $lr = 0.0005$ in our experiments, and use the exponential learning rate decay of 0.95 per step.
For regularization, we apply batch normalization~\cite{Sergey-icml-2015} to every dense layer, and use dropout~\cite{srivastava-jmlr-2014} after dense layers.
Our best model does not require any external data or pre-training other than Fashion IQ dataset. However, in order to improve ensemble performance, pretraining with external data was performed on variant models without attribute experts.
In the pre-training phase, we train each pre-training dataset in 3 epochs. And then we fine-tune the model with the fashion-IQ training dataset.  

In the test phase, we passed the sweeper filter for all possible target candidates. After that, calculating ranking using the distance with the Delivery Filter value

\section{Experiments}
\label{sec:experiments}

We report the experimental results of CurlingNet model for the fashion-IQ challenge.
The challenge provides an image retrieval dataset where the input query is specified in the form of a candidate image and two natural language expressions that describe the visual differences of the search target.
The dataset contains 77,683 fashion images and 30,134 pair of relative captions.
We strictly follow the evaluation protocols of the challenge.

\textbf{Evaluation Metrics.}
Every participant is evaluated by the recall metrics on the test splits of the dataset. 
For each of the three fashion categories (dresses, tops, shirts), Recall@10 and Recall@50 are computed on all test queries.
The overall performance is evaluated based on the average of Recall@10 and Recall@50. 

We defer more details and challenge rules to the challenge homepage\footnote{\url{https://sites.google.com/view/lingir/fashion-iq}.}.

\subsection{Quantitative Results}
Table~\ref{tbl:results}--\ref{tbl:results_ensemble} summarize the quantitative results of our experiments. 
Table~\ref{tbl:results} show the public validation results on single (\ie non-ensemble) models.
They share the same training pipeline for fair comparison.
For the Fashion-IQ~\cite{fashioniq-arxiv-2019} challenge, we compare with the results on the test dataset in Table~\ref{tbl:results_ensemble}, reported in the official evaluation server of Fashion-IQ as of the submission deadline (\ie Sep 30th, 2019 UTC 23:59).
We use the corresponding IDs and Team Name in the leaderboard to denote participants. 
In addition to the competitors in the leaderboard, we add a simple ablation baseline (Curling-concat), which conducts the simple fusion methods on a pair of source image and text query encoders. 
We also report the performance of state-of-the art models, FiLM~\cite{perez-AAAI-2018} and TIRG~\cite{vo-cvpr-2019} that exactly follow our setting of training environment, using the source codes provided by the original authors. 

Table \ref{tbl:results} compares between different methods on the validation dataset.
Our CurlingNet achieves the best retrieval performance with significant margins over the official baselines. 
The SUM baseline is based on the official start code, but shares the same image and text encoder setting with our model, trained with a triplet loss between (source image encoder + text encoder) and (target image encoder).
FiLM and TiRG model follows the setting in \cite{vo-cvpr-2019}. 
For Dual Encoder, we use their official code to train the model.

\subsection{Qualitative Results}
\begin{figure}[t]
\centering
\includegraphics[trim=0.0cm 0.0cm 0cm 0.0cm,clip,width=0.47\textwidth]{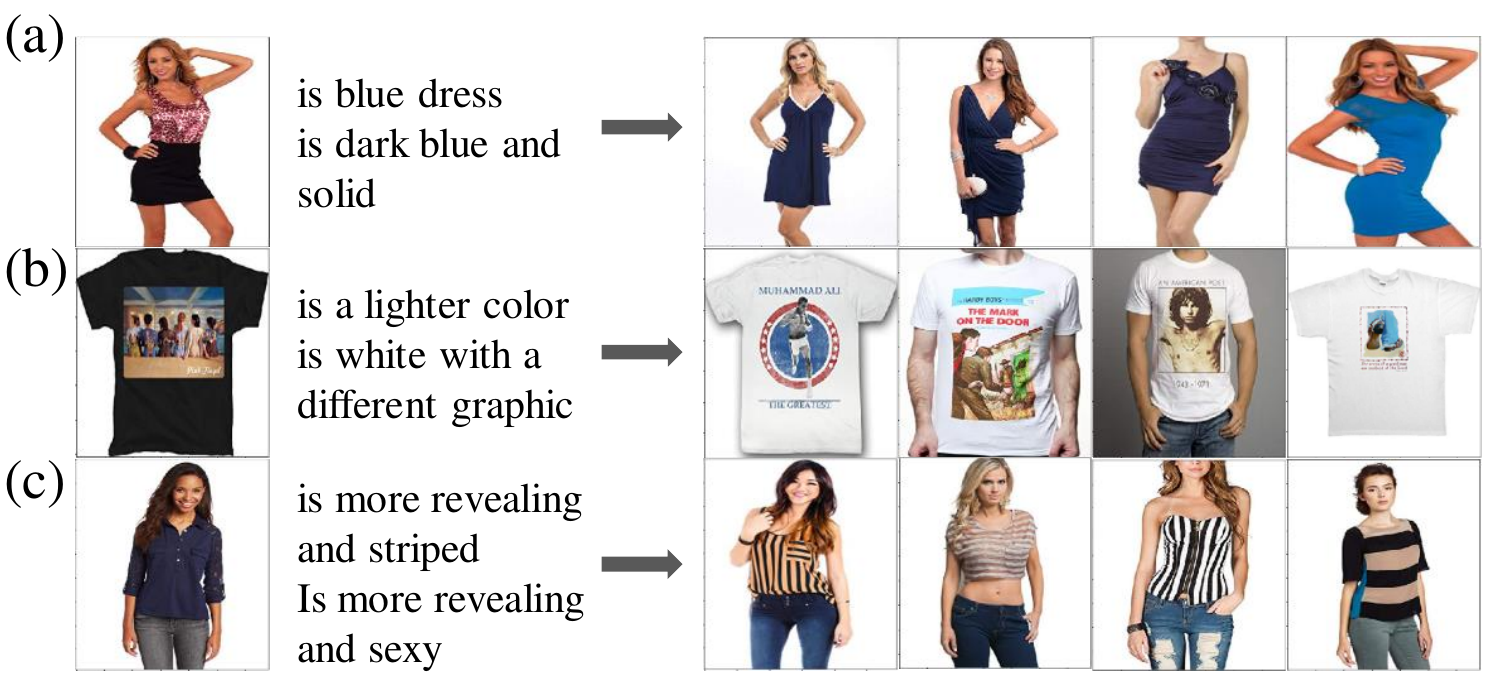}
\caption{Qualitative examples of the image-text retrieval task in the Fashion-IQ dataset.
}
\label{fig:examples}
\end{figure}

Figure \ref{fig:examples} illustrates qualitative results of our CurlingNet method.
We display source images and text queries (Left) and the highest-scored retrieved images (Right) for some test examples.
While maintaining the style of the source image, our model assigns a high score to the sample that well reflects the needs of the user query.

\begin{table*}
\begin{center}
\begin{tabular}{|l|cc|cc|cc|c|}
\hline
\multicolumn{1}{|c|}{Category} & \multicolumn{2}{c|}{Dress} & \multicolumn{2}{c|}{Shirt} & \multicolumn{2}{c|}{Toptee} & Total \\ 
\hline
\multicolumn{1}{|c|}{Metric}   & R@10         & R@50        & R@10        & R@50        & R@10         & R@50         & Avg   \\ 
\hline\hline
SUM   & 0.1120        & 0.2885       & 0.0873       & 0.2350       & 0.1086        & 0.2799        & 0.1852 \\
FiLM~\cite{perez-AAAI-2018}   & 0.1502        & 0.3748       & 0.1118       & 0.3076       & 0.1657        & 0.3941        &  0.2507 \\
Dual Encoder & 0.1720        & 0.4025       & 0.1398       & 0.3287       & 0.1815        & 0.3952        & 0.2699 \\ 
TiRG~\cite{vo-cvpr-2019}   & 0.1814        & 0.4253       & 0.1413       & 0.3415       & 0.1856        & 0.4171        &  0.2820 \\
\hline
Only-Resnet-Encoder (Pretrain)  & 0.2419        & 0.4863       & 0.1761       & 0.4038       & 0.2345        & 0.5109        & 0.3423  \\
Ours    & 0.2444        & 0.4769       & 0.1859       & 0.4057       & 0.2519        & 0.4966        & 0.3436 \\
Ensemble   & 	0.3302  & 	0.6009  &   0.2591 & 	0.5020 & 	0.3391  & 	0.6298 &   0.4435\\
\hline
\end{tabular}
\end{center}
\caption{Performance comparison for the Fashion-IQ dataset validation split in terms fo Recall@10 and Recall@50 (higher is better). Note that our model does not use external datasets.}
\label{tbl:results}
\end{table*}

\section{Conclusion}

We proposed the CurlingNet model for learning differential relations between two image embeddings with respect to text query.
The two key components of the model, Delivery and Sweeping filters, are easily adaptable in many deep metric learning tasks, including user adjustable retrieval or image recommendation systems.
We demonstrated that our method significantly improved the performance of image-text composition learning.
Our method achieved the top performance in Fashion-IQ challenge, and outperformed many state-of-the-art models for image-text retrieval tasks.
Moving forward, we plan to expand the applicability of the CurlingNet model;
since our method is applicable to any visual-text pair data,
we can explore other retrieval tasks on huge web data or multimodal conversational systems.

\begin{table}
\begin{center}
\begin{tabular}{|l|l|c|}
\hline
Submission       & Team Name  &{\footnotesize Avg R@(10,50)}   \\ \hline\hline
IvonaTau          &            & 0.1083    \\
NWPU-ASGO         &            & 0.1829     \\
Recmoon           &            & 0.1833   \\
nuanyang          &            & 0.2202   \\
abhinavagarwalla  &            & 0.2620    \\
cuberick          & CARDINAL   & 0.3289   \\
emd               &            & 0.3886   \\
skywalker         & DESIGNOVEL & 0.4367   \\
superraptors      &            & 0.4853   \\ \hline
Ours              & RippleAI   & 0.4680  \\
\hline
\end{tabular}
\end{center}
\medskip
\caption{
Performance comparison for final submission on Test Phase. 
We finished the second place in the challenge.
}
\vspace{-3pt}

\label{tbl:results_ensemble}
\end{table}

{\small
\bibliographystyle{ieee_fullname}
\bibliography{egbib}

\begin{thebibliography}{10}\itemsep=-1pt

\bibitem{arand-cvpr-2016}
Relja Arandjelovic, Petr Gronat, Akihiko Torii, Tomas Pajdla, and Josef Sivic.
\newblock Netvlad: Cnn architecture for weakly supervised place recognition.
\newblock In {\em CVPR}, 2016.

\bibitem{mutan-iccv-2017}
Hedi Ben-Younes, R{\'e}mi Cadene, Matthieu Cord, and Nicolas Thome.
\newblock Mutan: Multimodal tucker fusion for visual question answering.
\newblock In {\em ICCV}, 2017.

\bibitem{dual-cvpr-2019}
Jianfeng Dong, Xirong Li, Chaoxi Xu, Shouling Ji, Yuan He, Gang Yang, and Xun
  Wang.
\newblock Dual encoding for zero-example video retrieval.
\newblock In {\em CVPR}, 2019.

\bibitem{mcb-emnlp-2016}
Akira Fukui, Dong~Huk Park, Daylen Yang, Anna Rohrbach, Trevor Darrell, and
  Marcus Rohrbach.
\newblock Multimodal compact bilinear pooling for visual question answering and
  visual grounding.
\newblock In {\em EMNLP}, 2016.

\bibitem{fashioniq-arxiv-2019}
Xiaoxiao Guo, Hui Wu, Yupeng Gao, Steven Rennie, and Rogerio Feris.
\newblock The fashion iq dataset: Retrieving images by combining side
  information and relative natural language feedback.
\newblock In {\em ICCV}, 2019.

\bibitem{han-iccv-2017}
Xintong Han, Zuxuan Wu, Phoenix~X. Huang, Xiao Zhang, Menglong Zhu, Yuan Li,
  Yang Zhao, and Larry~S. Davis.
\newblock Automatic spatially-aware fashion concept discovery.
\newblock In {\em ICCV}, 2017.

\bibitem{he-arxiv-2015}
Kaiming He, Xiangyu Zhang, Shaoqing Ren, and Jian Sun.
\newblock {Deep Residual Learning for Image Recognition}.
\newblock In {\em CVPR}, 2016.

\bibitem{huang-cvpr-2017}
Gao Huang, Zhuang Liu, Laurens Van Der~Maaten, and Kilian~Q Weinberger.
\newblock Densely connected convolutional networks.
\newblock In {\em CVPR}, 2017.

\bibitem{Sergey-icml-2015}
Sergey Ioffe and Christian Szegedy.
\newblock {Batch Normalization: Accelerating Deep Network Training by Reducing
  Internal Covariate Shift}.
\newblock In {\em ICML}, 2015.

\bibitem{kingma-iclr-2015}
Diederik Kingma and Jimmy Ba.
\newblock {Adam: A Method for Stochastic Optimization}.
\newblock In {\em ICLR}, 2015.

\bibitem{Liu-bmvc-2019}
Y. Liu, S. Albanie, A. Nagrani, and A. Zisserman.
\newblock Use what you have: Video retrieval using representations from
  collaborative experts.
\newblock In {\em BMVC}, 2019.

\bibitem{miech-arxiv-2017}
Antoine Miech, Ivan Laptev, and Josef Sivic.
\newblock Learnable pooling with context gating for video classification.
\newblock {\em arXiv}, 2017.

\bibitem{Tomas-nips-2013}
Tomas Mikolov, Ilya Sutskever, Kai Chen, Greg Corrado, and Jeffrey Dean.
\newblock {Distributed Representations of Words and Phrases and their
  Compositionality}.
\newblock In {\em NIPS}, 2013.

\bibitem{perez-AAAI-2018}
Ethan Perez, Florian Strub, Harm De~Vries, Vincent Dumoulin, and Aaron
  Courville.
\newblock Film: Visual reasoning with a general conditioning layer.
\newblock In {\em AAAI}, 2018.

\bibitem{arxiv-fashiongen-2018}
N. {Rostamzadeh}, S. {Hosseini}, T. {Boquet}, W. {Stokowiec}, Y. {Zhang}, C.
  {Jauvin}, and C. {Pal}.
\newblock {Fashion-Gen: The Generative Fashion Dataset and Challenge}.
\newblock In {\em ArXiv}, 2019.

\bibitem{srivastava-jmlr-2014}
Nitish Srivastava, Geoffrey Hinton, Alex Krizhevsky, Ilya Sutskever, and Ruslan
  Salakhutdinov.
\newblock {Dropout: A Simple Way to Prevent Neural Networks from Overfitting}.
\newblock {\em JMLR}, 2014.

\bibitem{vo-cvpr-2019}
Nam Vo, Lu Jiang, Chen Sun, Kevin Murphy, Li-Jia Li, Li Fei-Fei, and James
  Hays.
\newblock Composing text and image for image retrieval-an empirical odyssey.
\newblock In {\em CVPR}, 2019.

\bibitem{amloss-arxiv-2019}
Yinfei Yang, Gustavo~Hernandez Abrego, Steve Yuan, Mandy Guo, Qinlan Shen,
  Daniel Cer, Yun-hsuan Sung, Brian Strope, and Ray Kurzweil.
\newblock Improving multilingual sentence embedding using bi-directional dual
  encoder with additive margin softmax.
\newblock In {\em arXiv}, 2019.

\end{thebibliography}
}

\end{document}